\documentclass[runningheads]{llncs}
\usepackage{graphicx}
\usepackage{cite}
\usepackage{amsmath,amssymb,amsfonts}
\usepackage{algorithmic}
\usepackage{graphicx}

\usepackage{xcolor}
\usepackage[caption=false]{subfig}
\usepackage{mathrsfs} 
\usepackage{babel}
\usepackage{fancyhdr}
\usepackage{geometry}
\usepackage{graphicx}
\usepackage{amssymb}
\usepackage{eucal}
\usepackage{setspace}

\newcommand{\beginsupplement}{%
        \setcounter{table}{0}
        \setcounter{figure}{0}
     }
\usepackage{xr}

\usepackage{xcolor}

\begin{document}
\title{BioLCNet: Reward-modulated Locally Connected Spiking Neural Networks}
%
%
\author{Hafez Ghaemi\textsuperscript{\textsection}\inst{1} \ \and
Erfan Mirzaei\textsuperscript{\textsection}\inst{2} \and
Mahbod Nouri\textsuperscript{\textsection}\inst{3} \and
Saeed Reza Kheradpisheh\textsuperscript{\dag}\inst{4}}
%
\authorrunning{H. Ghaemi, E. Mirzaei, M. Nouri, and S.R. Kheradpisheh}
%
\institute{Politecnico di Torino, Italy, \email{hafez.ghaemi@studenti.polito.it} \and
University of Tehran, Iran, \email{erfunmirzaei@ut.ac.ir} \and
The University of Edinburgh, United Kingdom, \email{m.nouri@sms.ed.ac.uk} \and
Shahid Beheshti University, Tehran, Iran, \email{s\_kheradpisheh@sbu.ac.ir}
}
%
\maketitle              

\begingroup\renewcommand\thefootnote{\textsection}
\footnotetext{Equal contribution}
\endgroup

\begingroup\renewcommand\thefootnote{\dag}
\footnotetext{Corresponding author}
\endgroup

\begin{abstract}
Brain-inspired computation and information processing alongside compatibility with neuromorphic hardware have made spiking neural networks (SNN) a promising method for solving learning tasks in machine learning (ML). Spiking neurons are only one of the requirements for building a bio-plausible learning model. Network architecture and learning rules are other important factors to consider when developing such artificial agents. In this work, inspired by the human visual pathway and the role of dopamine in learning, we propose a reward-modulated locally connected spiking neural network, BioLCNet, for visual learning tasks. To extract visual features from Poisson-distributed spike trains, we used local filters that are more analogous to the biological visual system compared to convolutional filters with weight sharing. In the decoding layer, we applied a spike population-based voting scheme to determine the decision of the network. We employed Spike-timing-dependent plasticity (STDP) for learning the visual features, and its reward-modulated variant (R-STDP) for training the decoder based on the reward or punishment feedback signal. For evaluation, we first assessed the robustness of our rewarding mechanism to varying target responses in a classical conditioning experiment. Afterwards, we evaluated the performance of our network on image classification tasks of MNIST and XOR MNIST datasets.
\end{abstract}


\section{Introduction}
\label{intro}
Deep convolutional neural network (DCNN) has been one of the best-performing architectures in the field of computer vision and object recognition \cite{Goodfellow-et-al-2016,lecun2015deep}. Despite the emergence of novel methods and architectures, such as visual transformers \cite{carion2020end}, DCNNs are still ubiquitous, and one of the most popular architectures employed for solving machine learning tasks. In addition, the representations formed in convolutional networks are similar to those found in the primate visual cortex \cite{schrimpf2020brain}. Nevertheless, traditional DCNN has fundamental differences with biological visual system.

First of all, neuron activations in an artificial neural network (ANN) are static real-numbered values that are modeled by differentiable, non-linear activation functions. This is in contrast to biological neurons that use discrete, and mostly sparse spike trains to transmit information between each other, and in addition to the rate of spikes (spatial encoding), they also use spike timing to encode information temporally \cite{tavanaei2019deep}. Therefore, a spiking neural network (SNN) is more akin to the neural networks in the brain. Spiking neural networks are more energy-efficient and compatible with neuromorphic hardware \cite{cao2015spiking}.

Secondly, the brain is incapable of error backpropagation, as done in traditional ANNs. One issue with error backpropagation in ANNs is the weight transport problem, i.e., the fact that weight connectivity in feedforward and feedback directions is symmetric \cite{liao2016important,NEURIPS2018_63c3ddcc}. Additionally, error feedback propagation, which does not affect neural activity, is not compliant with the feedback mechanisms that biological neurons use for communication \cite{lillicrap2020backpropagation}.

Furthermore, although convolutional neural network has shown great potential in solving any translation-invariant task, its use of weight sharing is biologically implausible. There is no empirical support for explicit weight sharing in the brain \cite{pogodin2021towards}. However, local connections between neurons are biologically plausible because neurons in the biological visual system exploit them to have local visual receptive fields \cite{gregor2010emergence}. Also, Poggio et al. \cite{poggio2017and} showed that DCNN can avoid the curse of dimensionality for compositional functions not due to its weight sharing, but its depth.

One of the main challenges in designing SNNs is learning from labeled data in a bio-plausible manner. To address this challenge and to enable the development of networks with deeper architectures, we decompose feature extraction from classification. We consider the classification task as a decision making problem, and propose a dopamine-based learning scheme.

Considering the above-mentioned arguments, and to move towards more bio-plausible architectures and learning methods in SNNs, in this work, we are proposing BioLCNet, a reward-modulated locally connected spiking neural network. Local filters enable the extension of the famous Diehl and Cook's architecture \cite{diehl2015unsupervised} into a more realistic model of the biological receptive fields in the visual pathway of the brain. We extended this locally-connected structure with a fully-connected layer to mimic the hierarchical structure of the visual pathway and to decompose the tasks of feature extraction and recognition. Our network is trained using the unsupervised spike-timing-dependent plasticity and its semi-supervised variant, reward-modulated STDP (R-STDP). The input images are encoded proportional to the pixels intensity using Poisson spike-coding that converts intensity to average neuron firing rate in Hertz. The hidden layer extracts features using unsupervised STDP learning. In the output layer, there are neuronal groups for each class label, trained using R-STDP, and decision making is based on aggregated number of spikes during the decision period. In this layer, in addition to normal R-STDP, we employed TD-STDP in SNNs \cite{fremaux2016neuromodulated} that is inspired by the dopamine hypothesis in the brain \cite{schultz1997neural} and incorporates the element of surprise in STDP learning. After conducting a classical conditioning experiment to prove the effectiveness of our decoding scheme and rewarding mechanism, we evaluate the classification performance of our network with different sets of hyperparameters on MNIST \cite{lecun1999mnist} and XOR MNIST datasets.

\section{Related Work}
\label{rel_work}
Neuroscientists and deep learning researchers have long been searching for more biologically plausible deep learning approaches in terms of neuronal characteristics, learning rules, and connection types. Regarding neuronal characteristics, researchers have turned to biological neuronal models and spiking neural networks. The vanishing performance gap between deep neural networks (DNNs) and SNNs, and the compatibility of SNNs with neuromorphic hardware and online on-chip training \cite{schemmel2010wafer} has piqued the interest of researchers \cite{mozafari2019bio}.

Spiking neurons are activated by discrete input spike trains. This differs from artificial neurons used in an ANN that have differentiable activation functions and can easily employ backpropagation and gradient-based optimization. There are works that use gradient-based methods with SNNs and some of them have achieved great performances \cite{kheradpisheh2020temporal,bellec2020solution}. On the other hand, many works in this area use derivations of the Hebbian learning rule where changes in connection weights depend on the activities of the pre and post-synaptic neurons. Spike-timing-dependent plasticity (STDP) and its variants, apply asymmetric weight updates based on the temporal activities of neurons. Normal STDP requires an external read-out for classification \cite{mozafari2018first}, and have been applied to image reconstruction and classification tasks by many researchers \cite{allred2016unsupervised,kheradpisheh2016bio,kheradpisheh2018stdp}. Reward-modulated STDP (R-STDP) uses a reward (or punishment) signal to directly modulate the STDP weight change, and can be used to decode the output without an external cue. Izhikevich \cite{izhikevich2007solving} tried to address the distal reward problem in reinforcement learning by using a version of R-STDP with decaying eligibility traces that gives recent spiking activity more importance. Around the same time, Florian \cite{florian2007reinforcement} showed that R-STDP can be employed to solve a simple XOR task with both Poisson and temporal encoding of the output. Historically, R-STDP was first adopted with temporal (rank-order) encoding for image classification \cite{mozafari2018first}. In this work, Mozafari et al. employed a convolutional architecture that uses a time-to-first-spike decoding scheme. An extended architecture was later developed which had multiple hidden layers \cite{mozafari2019bio}. The use of R-STDP with Poisson-distributed spike-coding has been mostly limited to fully-connected architectures for solving reinforcement learning robot navigation tasks \cite{bing2019end}. In a recent work, Weidel et al. \cite{weidel2021unsupervised} proposed a spiking network with clustered connectivity and R-STDP learning to solve multiple tasks including classification of a small subset of MNIST with three classes.

Despite the biological nature of local connections, they mostly underperform convolution-based methods with weight sharing in the visual domain, especially on large-scale datasets \cite{NEURIPS2018_63c3ddcc}. This weaker performance may be mainly attributed to the smaller number of parameters and better generalization in CNNs. Fewer parameters in CNNs would also require less memory and computational cost, and would lead to faster training \cite{poggio2017and}. Studies are being done to bridge the performance gap between convolutional and locally-connected networks \cite{lillicrap2020backpropagation,NEURIPS2018_63c3ddcc}.

The most prevalent architectures used for image classification in deep learning with both DNNs and SNNs are based on convolutional layers and weight sharing. However, as mentioned before, there are arguments against the biological plausibility of these approaches \cite{NEURIPS2018_63c3ddcc,pogodin2021towards}. Locally connected (LC) networks are an alternative to the convolutional ones. Illing et al. \cite{illing2019biologically} show that shallow networks with localized connectivity and receptive fields perform much better than fully-connected networks on the MNIST benchmark. However, Bartunov et al. \cite{NEURIPS2018_63c3ddcc} showed that the lower generalization of LC networks compared to CNNs results in their underperforming CNNs in most image classification tasks, and prevents their scalability to larger datasets such as ImageNet. Recently, Pogodin et al. \cite{pogodin2021towards} proposed bio-inspired dynamic weight sharing and adding lateral connections to locally-connected layers to achieve the same regularization goals of weight sharing and normal convolutional filters. Saunders et al. \cite{saunders2019locally} and Illing et al. \cite{illing2019biologically} used local filters for learning visual feature representations in SNNs, and achieved a good performance on the MNIST benchmark with supervised decoding mechanisms. 


\section{Theory}
\label{theory}
In this section, we will outline the theoretical foundations underlying our proposed method. Specifically, the dynamics of the spiking neuronal model, the learning rules used, and the connection type employed in our network will be described.
\subsection{Adaptive LIF neuron model}
The famous leaky and integrate fire neuronal model is governed by the following differential equation \cite{gerstner2014neuronal},

\begin{equation}\label{eq1}
\tau_m \frac{du}{dt} = -[u(t)-u_{rest}]+RI(t),
\end{equation}

where $u(t)$ denotes the neuron membrane potential and is a function of time, $R$ is the membrane resistance, $I(t)$ is any arbitrary input current, and $\tau_m$ is the membrane time constant. Equation~(\ref{eq1}) dictates that the neuron potential exponentially decays to a constant value $u_{rest}$ over time. When a pre-synaptic neuron fires (spikes), it generates a current that reaches its post-synaptic neurons. In the simple leaky integrate and fire (LIF) model, a neuron fires when its potential surpasses a \textbf{constant} threshold $u_{thr}$. After firing, the neuron's potential resets to a constant $u_{reset}$ and will not be affected by any input current for a period of time known as the refractory period ($\Delta t_{ref}$).

A variant of the LIF model uses adaptive firing thresholds. In this model, $u_{thr}$ can change over time based on the neuron's rate of activity \cite{diehl2015unsupervised}. When a neuron fires, its tolerance to the input stimuli and consequently its firing threshold increases by a constant amount, $g_0$, otherwise the threshold decays exponentially with a time constant $\tau_g$ to the default threshold $u_{thr_0}$. Equations (\ref{eq2}) to (\ref{eq4}) explain the dynamics of the adaptive LIF model,
\begin{equation}\label{eq2}
u_{thr}(t) = u_{thr_0} + g(t),
\end{equation}
where,
\begin{equation}\label{eq3}
\tau_g d_g/d_t = -g(t),
\end{equation}
and
\begin{equation} \label{eq4}
spike \Rightarrow g(t) = g(t-1) + g_0,
\end{equation}

\subsection{Reward-modulated STDP}
Spike-timing-dependent plasticity is a type of biological Hebbian learning rule that is also aligned with human intuition ("Neurons that fire together wire together." \cite{lowel1992selection}). The normal STDP is characterized by two asymmetric update rules. The synaptic weights are updated based on the temporal activities of pre and post-synaptic neurons. When a pre-synaptic neuron fires shortly \textbf{before} its post-synaptic neuron, the causal connection between the first and the second neuron temporal activity is acknowledged, and the connection weight is increased. On the other hand, if the post-synaptic neuron fires shortly \textbf{after} the pre-synaptic neuron, the causality is undermined and the synaptic strength will decrease \cite{hebb1949organisation}. These weight updates, called long-term potentiation (LTP) and long-term depression (LTD), can be performed with asymmetric learning rates to adapt the learning rule to the excitatory-to-inhibitory neuron ratio or the connection patterns of a specific neural network. A popular variant of STDP that integrates reinforcement learning into the learning mechanism of spiking neural networks is reward-modulated STDP (also known as R-STDP or MSTDP \cite{florian2007reinforcement}). In R-STDP, a global reward or punishment signal, which can be a function of time, is generated as the result of the network's activity or task performance. Using a notation similar to Florian \cite{florian2007reinforcement}, to mathematically formulate both STDP and R-STDP, we can define the spike train of a pre-synaptic neuron as the sum of Dirac functions over the spikes of the post-synaptic neurons,

\begin{equation} \label{eq5}
\Phi(t) = \sum\mathop{}_{\mathscr{F}_i} \delta(t-t_i^f).
\end{equation}

where $t_i^f$ is the firing time of the $i^{th}$ post-syanptic neuron. Now, we can define the variables $P_{ij}^+$ and $P_{ij}^-$ to respectively track the influence of pre or post-synaptic spikes on weight updates. Now, the spike trace $\xi$ for a given spike from neuron $i$ to $j$ can be defined as below,
\begin{equation} \label{eq6}
\xi_{ij}=P_{ij}^+\Phi_i(t)+P_{ij}^-\Phi_j(t),
\end{equation}
where,
\begin{equation} \label{eq7}
dP_{j}^+/dt=-P_{j}^+/\tau_{+}+\eta_{post}\Phi_j(t),
\end{equation}
\begin{equation} \label{eq8}
dP_{i}^-/dt=-P_{i}^-/\tau_{-}-\eta_{pre}\Phi_i(t),
\end{equation}
where we assumed that $P_{ij} = P_{j}$ for all pre-synaptic connections related to neuron $j$, and $P_{ij} = P_{i}$ for all post-synaptic connections related to neuron $i$.

The variables $\tau_{\pm}$ are the time constants determining the time window in which a spike can affect the weight updates. Using larger time constants will cause spikes that are further apart to also trigger weight updates. The variables $\eta_{post}$ and $\eta_{pre}$ determine the learning rate for LTP and LTD updates respectively. We denote the reward or punishment signal with $r(t)$. The R-STDP update rules for positive and negative rewards can be written as,

\begin{equation} \label{eq9}
\frac{dw_{ij}(t)}{dt}=\gamma r(t) \xi_{ij}(t),
\end{equation}

where $\gamma$ is a scaling factor. The update rule for normal STDP can also be written as,

\begin{equation} \label{eq10}
\frac{dw_{ij}(t)}{dt}=\gamma \xi_{ij}(t).
\end{equation}

Based on Equation~(\ref{eq9}), we note that R-STDP updates only take effect when a non-zero modulation signal is received at time step $t$. However, STDP updates do not depend on the modulation signal, and are applied at every time step. In other words, STDP can be considered a special case of R-STDP where the reward function is equal to 1 in every time step. This causes STDP to respond to the most frequent patterns regardless of their desirability.

\subsection{Local connections}
\begin{figure}[h]
\begin{center}
\includegraphics[width=0.6\textwidth]{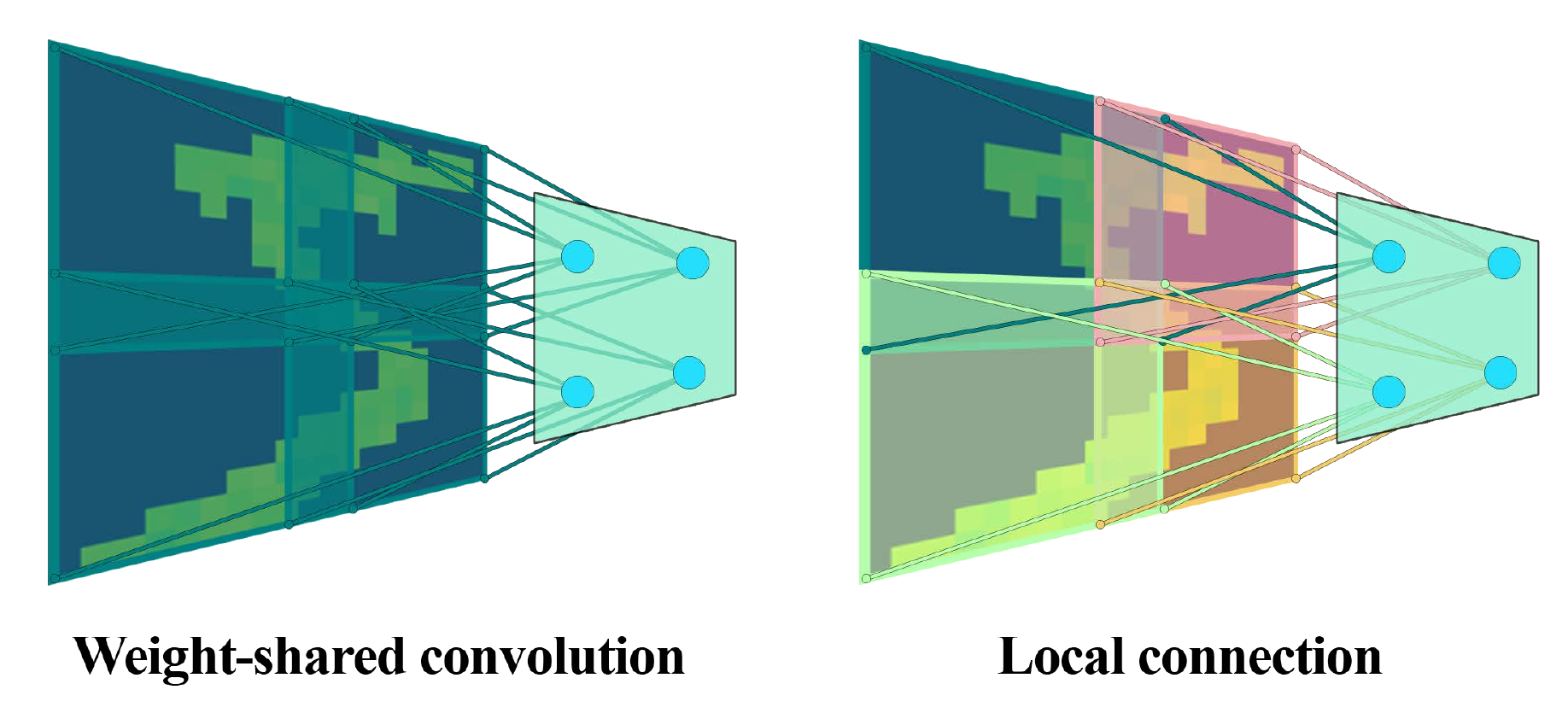}
\caption{Visual comparison of convolutional and local connections for a given channel; in convolutional connections, the weights are shared between all receptive fields. However, in a local connections, each receptive field has its own set of weights.}
\label{fig1}
\end{center}
\end{figure}

A local connection in a neural network is similar to a convolutional connection but with distinct filters for each receptive field. As seen in Fig.~\ref{fig1}, in normal convolutional connections, there is one filter for each channel that is convolved with all receptive fields as it moves along the layer's input. This filter has one set of weights that are updated using the network's update rule. However, In local connection (LC), after taking each stride, a new set of parameters characterize a whole new filter for the next receptive field. This type of connectivity between the input and the LC layer resembles the physical structure of retinal Ganglion cells. Because there are more filters in an LC, the number of distinct synapses in a local connection is greater than a convolutional connection, yet much lower than a dense connection. Similar to a convolutional connection, assuming square filters, and equal horizontal and vertical strides, we can specify a local connection by the number of channels ($ch_{lc}$), the kernel size ($k$), and the stride ($s$). 

\section{Architecture and Methods}
\begin{figure}[h]
\begin{center}
\includegraphics[width=0.6\textwidth]{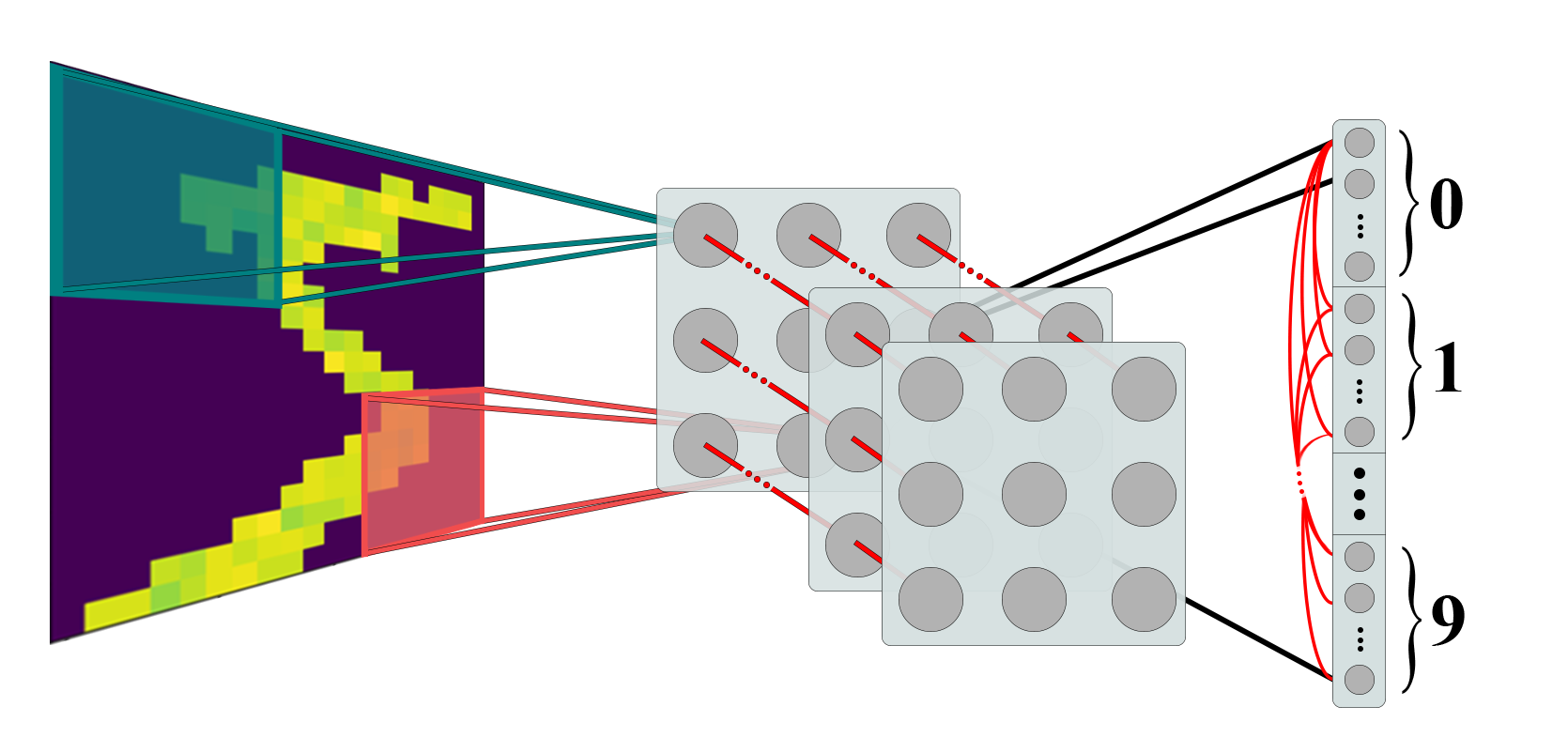}
\caption{Graphical representation of the proposed network; locally connected filters will be applied to the Poisson-distributed input image. Based on a winner-take-all inhibition mechanism, the most relevant features from each receptive field transmit their spikes to the decoding layer, which selects the most active neuronal group as the predicted label exploiting lateral inhibitory connections. The red lines indicate inhibitory connections.}
\label{fig2}
\end{center}
\end{figure}

BioLCNet consists of an input layer, a locally connected hidden layer, and a decoding layer. Each layer structure and its properties alongside the training and rewarding procedure will be delineated in this section. A graphical representation of our network is presented in Fig.~\ref{fig2}. During our experiments, the simulation time $T$ is divided into three phases, adaptation period ($T_{adapt}$), decision period ($T_{dec}$), and learning period ($T_{learn}$). The details of each phase will be specified in the remainder of this section.

\subsection{Encoding layer}
The input of the network is an image of dimensions ($ch_{in}$, $h_{in}$, $w_{in}$). For a grayscale image dataset such as MNIST, $ch_{in}$ equals to one. Each input channel is encoded using a Poisson-distributed scheme, i.e, the spiking neuron corresponding to each pixel has an average firing rate proportional to the intensity of that pixel. By choosing the maximum firing rate $f_{max}$, the spike trains average firing rates will be distributed in the interval [0, $f_{max}$] Hertz based on the pixel values.

\subsection{Feature extraction layer (local connections)}
The encoded input at each simulation time step passes through local connections with $ch_{out}$ distinct filters for each receptive field. Therefore, the output of this layer will have dimensions ($ch_{out}$, $h_{out}$, $w_{out}$), where the output size depends on the size of the kernel and the stride. There are generally two approaches in the SNN literature for training a feature extraction layer with Poisson-distributed inputs using STDP to attain a rich feature representation and also prevent the weights from growing too large. One is allowing the weights to have negative values, which corresponds to having inhibitory neurons, as done in the convolutional layers used by Lee et al. \cite{lee2018deep}. The other is to use a combination of recurrent inhibitory connections and adaptive thresholds \cite{diehl2015unsupervised,saunders2018stdp,saunders2019locally}. In this work, we used the latter approach for our feature extraction LC layer. We use adaptive LIF neurons and inhibitory connections between neurons that share the same receptive field. This is equivalent to the winner-take-all inhibition mechanism which causes a competition between neurons to select the most relevant features. The inhibitory connections are non-plastic and they all have a static negative weight $w_{inh}$ with a large absolute value. As hypothesized by Diehl and Cook \cite{diehl2015unsupervised}, the adaptive threshold of the neurons in this layer is a measure that may counterbalance the large number of inhibitory connections to each neuron, that is not compliant with the biological 4:1 excitatory to inhibitory ratio \cite{connors1990intrinsic}.

In normal STDP, the LTP learning rate ($\eta_{post}$) is usually chosen larger than the LTD rate ($\eta_{pre}$) to suppress the random firing of neurons that triggers many LTD updates during the early stages of training. However, this may become problematic in the later stages, and the weights may grow too large. Therefore, in practice, different mechanisms, such as weight clipping and normalization are used to prevent the weights running amok. In this work, we clipped the weights to stay in the range $[0, 1]$. We also employed the normalization technique \cite{diehl2015unsupervised,saunders2019locally} and normalized the pre-synaptic weights of each neuron in the LC layer to have a constant mean of $c_{norm}$ at the end of each time step.

\subsection{Decoding layer and rewarding mechanisms}
The final layer of our network is a fully connected layer for reward-based decoding. The layer is divided into $n_c$ neuronal groups where $n_c$ is the number of classes related to the task. Consequently, the $n_{out}$ neurons in this layer are divided equally into $n_c$ neuronal groups. The predicted label for a given test sample is the class whose group has the most number of spikes aggregated over the decision period ($T_{dec}$). This decoding layer is trained using reinforcement learning and R-STDP during the learning period ($T_{learn}$) based on the modulation signal generated by the rewarding mechanism. We exploited two different rewarding mechanisms, normal R-STDP \cite{florian2007reinforcement}, and TD-STDP \cite{fremaux2016neuromodulated}. In normal R-STDP, we use a fixed reward or punishment signal for the whole learning period ($T_{learn}$) based on the prediction of the network for the $i^{th}$ training sample,

\begin{equation} \label{eq11}
r_i =\left \{ \begin{array}{rcl}
1:
& predicted\ label = target\ label \\ -1: & otherwise
\end{array}\right.
\end{equation}

The second mechanism, TD-STDP is based on the reward prediction error theory in neuroscience and reinforcement learning. According to this theory, the dopaminergic neurons in the brain release dopamine proportional to the difference between the actual reward and the expected reward, not solely based on the actual reward \cite{schultz1997neural,sutton2018reinforcement}. This mechanism involves the element of surprise in learning; the agent receives an amplified reward signal when it has a correct prediction after a sequence of wrong ones. Similarly, it receives an amplified punishment signal when a wrong prediction comes after a sequence of correct ones. We can formulate TD-STDP with exponential moving average as below,

\begin{equation} \label{eq12}
M_i = \eta_{rpe}(r_{i}-\mathrm{EMA}_r)
\end{equation}

where $M_i$ is the scalar TD-STDP modulation signal used during the whole learning period ($T_{learn}$) of the $i^{th}$ training sample, $r_{i}$ is the reward signal received based on the prediction in~(\ref{eq11}), and $\mathrm{EMA}_r$ is the exponential moving average of the reward signals with a smoothing factor $\alpha$.

\begin{figure*}[htp]
    \centering
  \subfloat(a\label{3a}){%
       \includegraphics[width=0.27\linewidth]{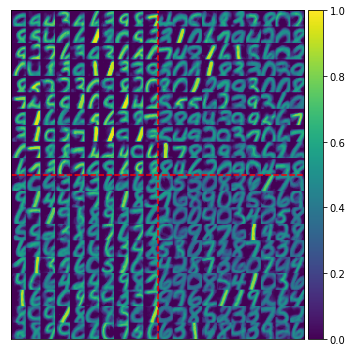}}
    \hfill
  \subfloat(b\label{3b}){%
        \includegraphics[width=0.27\linewidth]{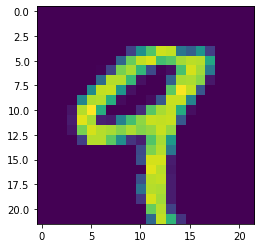}}
  \subfloat(c\label{3c}){%
        \includegraphics[width=0.27\linewidth]{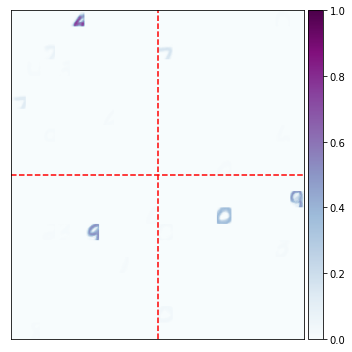}}
    \hfill
  \caption{Input and LC layer visualizations. (a) LC layer learned filters; the red lines separate filters corresponding to each receptive field. (b) A sample input image. (c) The LC layer activation map corresponding to the sample input image shown.}
  \label{fig3} 
\end{figure*}

\subsection{Training procedure}
The network is trained in a layer-wise fashion. After initializing the weights uniformly random between $[0,1]$, we train the feature extraction LC layer in a completely unsupervised manner using STDP. Simulation time for training the feature extraction layer is $T_{learn}$ time steps. After this layer is trained, the weights are freezed, and we train the decoding FC layer in a semi-supervised manner using R-STDP and the selected rewarding mechanism. Training this layer requires all three simulation phases. The input image is first presented to the network for $T_{adapt}$ time steps to let the LC layer neurons adapt to the input image and relevant features are selected based on the WTA mechanism. During $T_{dec}$ time steps, the decoding layer accumulates the number of spikes received by each neuronal group to determine the predicted label. It was observed that a small $T_{adapt}$ highly affected the quality of the network's decision. In other words, it is important to let the WTA mechanism take effect before entering the decision phase. Afterwards, the modulation signal is generated and the decoding layer weights are updated using R-STDP for a duration of $T_{learn}$ time steps. It should be emphasized that we do not use the ground truth label in any of the training steps, and the feedback signal is generated only based on the validity of the network's decision.

When training the LC layer, we observed that after a specific number of iterations (training samples), the weights of this layer converge and remain constant. Fig. 3a visualizes the filters learned after 2000 iterations for 100 channels with a filter (kernel) size of 15 with a stride of 4 applied to the input images of the MNIST dataset. This fast convergence is an evidence showing the strength of STDP learning. Considering these observations, and to save computation time, we limit the number of training sample of the LC layer to 2000 for all of the hyperparameter configurations for the two classification tasks. Given a sample MNIST input image (Fig. 3b), we plotted the activation map of the LC layer (Fig. 3c). This map shows that the post-synaptic neurons corresponding to the relevant features activate, and suppress the other neurons in accordance with the WTA inhibition mechanism.

The network is implemented using PyTorch \cite{NEURIPS2019_9015}, and mostly on top of the BindsNet framework \cite{hazan2018bindsnet}. We reimplemented the local connection topology to make it compatible with multi-channel inputs and a possible deep extension of our network. For transparency and to foster reproducibility, the code of the experiments are available publicly\footnote{\url{https://github.com/Singular-Brain/BioLCNet}}.

\begin{figure}[h] 
\begin{center}
\includegraphics[width=0.6\textwidth]{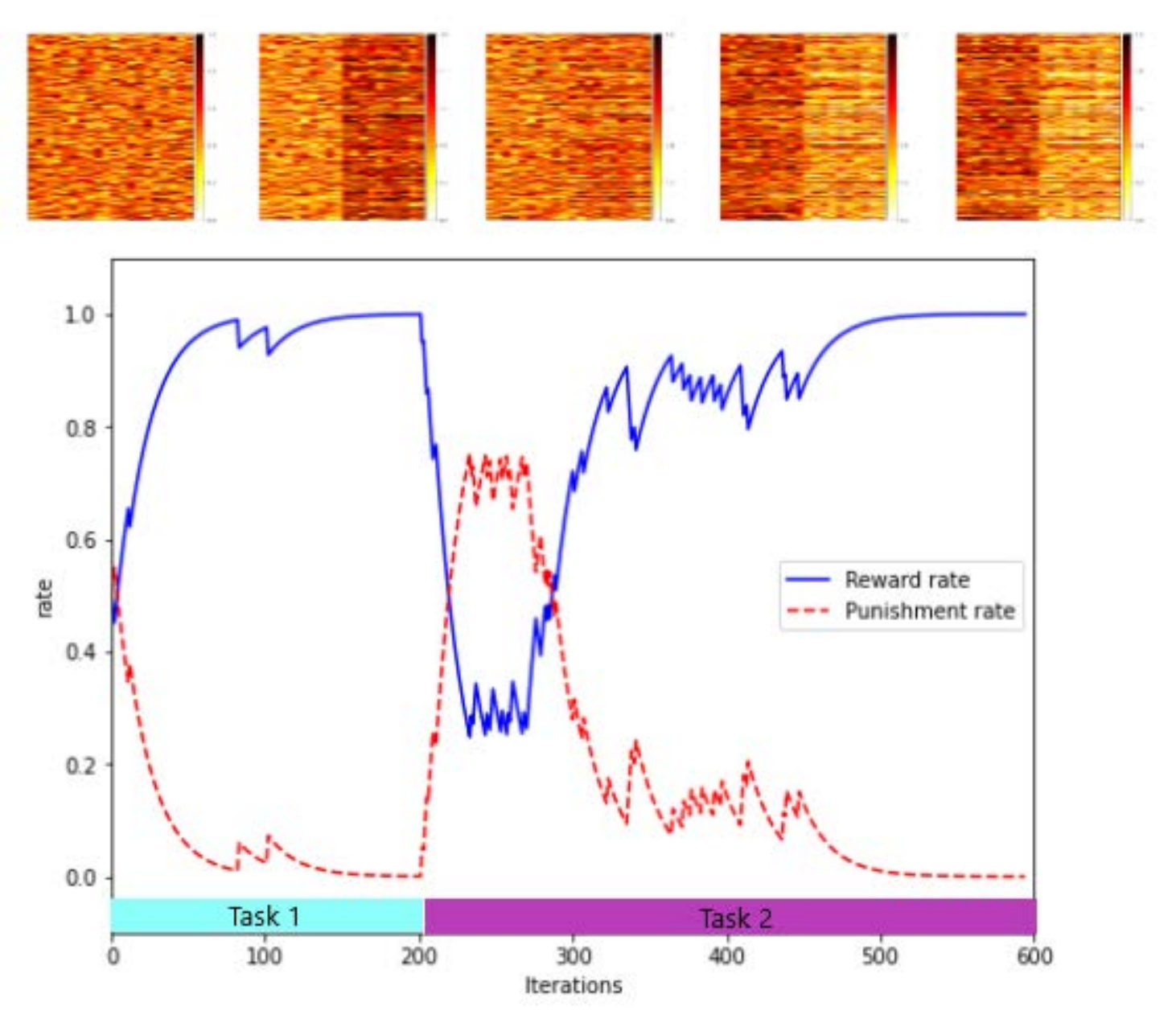}
\caption{Classical conditioning experiment; in this experiment, we tested the adaptability of the network to varying target responses. The plot shows the rate of receiving reward and punishment averaged over 20 runs, and the decoding layer weight maps at iterations 0, 200, 300, 400, and 600. The right half of the weight heat maps correspond to the task 1 target response neurons, and the left half corresponds to the task 2 target response neurons. The weights adapt to the varying target response during the experiment.}
\label{fig4}
\end{center}
\end{figure}

\section{Experiments and Discussion}
\subsection{Classical conditioning}
In order to show the effectiveness of our rewarding mechanism, we perform a classical (Pavlovian) conditioning experiment. This type of conditioning pairs up a neutral stimulus with an automatic conditioned response by the agent. In this experiment, we present the network with images belonging to one class of the MNIST dataset as the neutral stimuli. We used the pre-trained feature extraction layer of a model with 25 channels with filter size of 13 and stride of 3, following by a decoding layer with 20 neurons for a two-class prediction task. In the first half of the experiment (task 1), the target response is class 1, and the network receives a constant reward of 1 if it predicts this class regardless of the input. A punishment signal of -1 is received if the agent predicts class 0. We monitor the rate of the reward and punishment received during the experiment. After the convergence in about 50 iterations, Fig.~\ref{fig4} shows that the agent has become completely conditioned on the rewarding response. After 200 iterations, we swap the rewarding and punishing classes, and continue running the network. In task 2, the network should predict the input images as class 0. The RL agent (the network) adapts to the change notably fast, and completely changes its behavior after about 100 iterations. The heat maps in Fig.~\ref{fig4} visualize the weights of the output layer through the training.

The reward adaptability of an RL agent is critical because in many real-world problems the environment is non-stationary. Integration of reward adaptation into spiking neural networks, as done in this work, can pave the path for models that simulate human behaviour with the same spike-based computation as done in the human brain.

\subsection{MNIST}
To evaluate our network's classification performance, we trained our model on the MNIST benchmark \cite{lecun1999mnist}. Some of the hyperparameters were fixed and others were subject to grid search. The full list of hyperparameters for this experiment are given in Table~\ref{table1} in supplementary materials.

\begin{figure}[htbp]
\begin{center}
\includegraphics[width=0.65\textwidth]{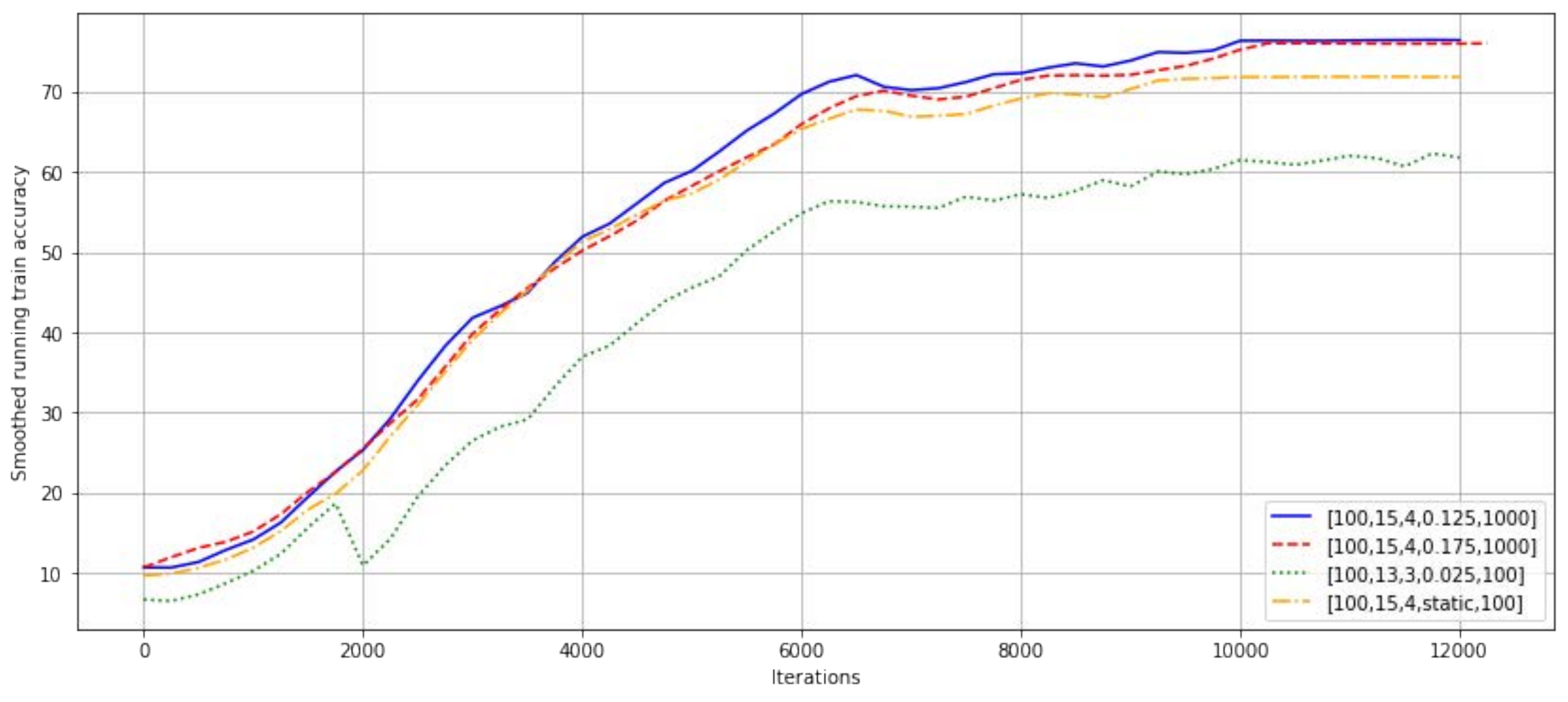}
\caption{Smoothed running accuracy over the training set for four sets of hyperparameters using the R-STDP classifier}
\label{fig5}
\end{center}
\end{figure}

Considering the hyperparameters mentioned in Table 1 of the supplementary materials, we report in Table~\ref{table1}, the classification accuracy on the whole MNIST test set (10000 samples) for four hyperparameter configurations chosen based on the highest test accuracy obtained after conducting a grid search. The number of neurons and synapses for each model are also reported in this table. The final models were all trained using 10000 training samples from the MNIST training set. Using more training samples did not improve the classification performance as can be observed from Fig.~\ref{fig5}. The mean and standard deviations reported are estimated from ten independent runs. In addition to the RL-based models, another classification approach was employed. In this approach, for each training sample, we create a feature vector containing the number of spikes aggregated over $T_{learn}$ time steps for every filter in the LC layer. We use these feature vectors to train a support vector machine (SVM) classifier. The SVM results are also obtained by training on 10000 training samples, and testing on the whole MNIST test set. The SVM test results for two different hyperparameter configurations are reported in Table~\ref{table1} and are compared to the RL-based results. The best performance of SVM and RL-based classification are 87.50, and 76.40 respectively.

\begin{table*}[htbp]
\begin{center}
\centering
\caption{MNIST test dataset accuracies obtained by four different sets of hyper-parameters; the test accuracies are averaged over ten independent runs}
\label{table1}
\small
\begin{tabular}{|c|c|c|c|c|}
\hline
Parameters [$k$, $s$, $\eta_{rpe}$, $n_{out}$]  & $n_{neurons}$ & $n_{synapses}$ & Test accuracy & SVM test accuracy\\ \hline
[13, 3, 0.025, 100]           & 1700 & 430400 &  61.30 $\pm$3.14 & 87.5$\pm$1.32       \\ \hline
[15, 4, 0.175, 1000]          & 1884 & 490000 &  75.00 $\pm$2.68  & 83.3$\pm$1.74    \\ \hline
[15, 4, 0.125, 1000]         & 1884 & 490000 &  76.40 $\pm$2.43 &  83.3$\pm$1.74   \\ \hline
[15, 4, (static), 100]           & 984 & 130000 &  68.8 $\pm$2.87   & 83.3$\pm$1.74    \\ \hline
\end{tabular}
\end{center}
\end{table*}

We observed that the features that are activated in the hidden layer and transferred to the decoding layer may overlap in samples with different labels but similar form, which may lower the classification performance of the RL-based and SVM classifiers. This claim is supported by the low SVM performance compared to its usual performance on the MNIST benchmark. Two important observations can be made from Table~\ref{table1}. First, the classification accuracy has a positive correlation with the filter size, and the number of neurons in the decoding layer. Consequently, training a larger network with higher number of channels and decoding neurons may be a possible way to improve the classification performance. This solution was neglected due to an exponential growth of the training time when increasing the number of channels. Secondly, as shown in Table~\ref{table1}, using TD-STDP instead of normal (static) R-STDP improved the classification performance. Indeed, in TD-STDP, the agent exploits the element of surprise to modulate its learning signal, which in turn, would result in a more stable learning. The results of another experiment on the XOR MNIST dataset is provided in the supplementary materials.

\section{Conclusions and Future Work}
In this work, we examined the capabilities of a neural network with three-fold biological plausibility; spiking neurons, local visual receptive fields, and a reward-modulated learning rule. In the future, by bringing ideas such as dynamic weight sharing and lateral connections \cite{pogodin2021towards} to spiking neural networks, we may be able to obtain richer feature representations using locally connected SNNs. Exploiting structures, such as pooling layers, along with advances in SNN minibatch processing \cite{saunders2020minibatch} and neuromorphic hardware \cite{schemmel2010wafer}, we can extend our network with larger and deeper architectures to solve more complex tasks.

\bibliographystyle{splncs04}
\bibliography{bib}

\begin{thebibliography}{10}
\providecommand{\url}[1]{\texttt{#1}}
\providecommand{\urlprefix}{URL }
\providecommand{\doi}[1]{https://doi.org/#1}

\bibitem{allred2016unsupervised}
Allred, J.M., Roy, K.: Unsupervised incremental stdp learning using forced
  firing of dormant or idle neurons. In: 2016 International Joint Conference on
  Neural Networks (IJCNN). pp. 2492--2499. IEEE (2016)

\bibitem{NEURIPS2018_63c3ddcc}
Bartunov, S., Santoro, A., Richards, B., Marris, L., Hinton, G.E., Lillicrap,
  T.: Assessing the scalability of biologically-motivated deep learning
  algorithms and architectures. In: Bengio, S., Wallach, H., Larochelle, H.,
  Grauman, K., Cesa-Bianchi, N., Garnett, R. (eds.) Advances in Neural
  Information Processing Systems. vol.~31. Curran Associates, Inc. (2018)

\bibitem{bellec2020solution}
Bellec, G., Scherr, F., Subramoney, A., Hajek, E., Salaj, D., Legenstein, R.,
  Maass, W.: A solution to the learning dilemma for recurrent networks of
  spiking neurons. Nature communications  \textbf{11}(1),  1--15 (2020)

\bibitem{bing2019end}
Bing, Z., Jiang, Z., Cheng, L., Cai, C., Huang, K., Knoll, A.: End to end
  learning of a multi-layered snn based on r-stdp for a target tracking
  snake-like robot. In: 2019 International Conference on Robotics and
  Automation (ICRA). pp. 9645--9651. IEEE (2019)

\bibitem{cao2015spiking}
Cao, Y., Chen, Y., Khosla, D.: Spiking deep convolutional neural networks for
  energy-efficient object recognition. International Journal of Computer Vision
   \textbf{113}(1),  54--66 (2015)

\bibitem{carion2020end}
Carion, N., Massa, F., Synnaeve, G., Usunier, N., Kirillov, A., Zagoruyko, S.:
  End-to-end object detection with transformers. In: European Conference on
  Computer Vision. pp. 213--229. Springer (2020)

\bibitem{connors1990intrinsic}
Connors, B.W., Gutnick, M.J.: Intrinsic firing patterns of diverse neocortical
  neurons. Trends in neurosciences  \textbf{13}(3),  99--104 (1990)

\bibitem{diehl2015unsupervised}
Diehl, P.U., Cook, M.: Unsupervised learning of digit recognition using
  spike-timing-dependent plasticity. Frontiers in computational neuroscience
  \textbf{9}, ~99 (2015)

\bibitem{florian2007reinforcement}
Florian, R.V.: Reinforcement learning through modulation of
  spike-timing-dependent synaptic plasticity. Neural computation
  \textbf{19}(6),  1468--1502 (2007)

\bibitem{fremaux2016neuromodulated}
Fr{\'e}maux, N., Gerstner, W.: Neuromodulated spike-timing-dependent
  plasticity, and theory of three-factor learning rules. Frontiers in neural
  circuits  \textbf{9}, ~85 (2016)

\bibitem{gerstner2014neuronal}
Gerstner, W., Kistler, W.M., Naud, R., Paninski, L.: Neuronal dynamics: From
  single neurons to networks and models of cognition. Cambridge University
  Press (2014)

\bibitem{Goodfellow-et-al-2016}
Goodfellow, I., Bengio, Y., Courville, A.: Deep Learning. MIT Press (2016),
  \url{http://www.deeplearningbook.org}

\bibitem{gregor2010emergence}
Gregor, K., LeCun, Y.: Emergence of complex-like cells in a temporal product
  network with local receptive fields (2010)

\bibitem{hazan2018bindsnet}
Hazan, H., Saunders, D.J., Khan, H., Patel, D., Sanghavi, D.T., Siegelmann,
  H.T., Kozma, R.: Bindsnet: A machine learning-oriented spiking neural
  networks library in python. Frontiers in neuroinformatics  \textbf{12}, ~89
  (2018)

\bibitem{hebb1949organisation}
Hebb, D.O.: The organisation of behaviour: a neuropsychological theory. Science
  Editions New York (1949)

\bibitem{illing2019biologically}
Illing, B., Gerstner, W., Brea, J.: Biologically plausible deep learning—but
  how far can we go with shallow networks? Neural Networks  \textbf{118},
  90--101 (2019)

\bibitem{izhikevich2007solving}
Izhikevich, E.M.: Solving the distal reward problem through linkage of stdp and
  dopamine signaling. Cerebral cortex  \textbf{17}(10),  2443--2452 (2007)

\bibitem{kheradpisheh2016bio}
Kheradpisheh, S.R., Ganjtabesh, M., Masquelier, T.: Bio-inspired unsupervised
  learning of visual features leads to robust invariant object recognition.
  Neurocomputing  \textbf{205},  382--392 (2016)

\bibitem{kheradpisheh2018stdp}
Kheradpisheh, S.R., Ganjtabesh, M., Thorpe, S.J., Masquelier, T.: Stdp-based
  spiking deep convolutional neural networks for object recognition. Neural
  Networks  \textbf{99},  56--67 (2018)

\bibitem{kheradpisheh2020temporal}
Kheradpisheh, S.R., Masquelier, T.: Temporal backpropagation for spiking neural
  networks with one spike per neuron. International Journal of Neural Systems
  \textbf{30}(06),  2050027 (2020)

\bibitem{lecun1999mnist}
LeCun, Y., Cortes, C., Burges, C.: The mnist dataset of handwritten digits
  (images). NYU: New York, NY, USA  (1999)

\bibitem{lecun2015deep}
LeCun, Y., Bengio, Y., Hinton, G.: Deep learning. nature  \textbf{521}(7553),
  436--444 (2015)

\bibitem{lee2018deep}
Lee, C., Srinivasan, G., Panda, P., Roy, K.: Deep spiking convolutional neural
  network trained with unsupervised spike-timing-dependent plasticity. IEEE
  Transactions on Cognitive and Developmental Systems  \textbf{11}(3),
  384--394 (2018)

\bibitem{liao2016important}
Liao, Q., Leibo, J., Poggio, T.: How important is weight symmetry in
  backpropagation? In: Proceedings of the AAAI Conference on Artificial
  Intelligence. vol.~30 (2016)

\bibitem{lillicrap2020backpropagation}
Lillicrap, T.P., Santoro, A., Marris, L., Akerman, C.J., Hinton, G.:
  Backpropagation and the brain. Nature Reviews Neuroscience  \textbf{21}(6),
  335--346 (2020)

\bibitem{lowel1992selection}
Lowel, S., Singer, W.: Selection of intrinsic horizontal connections in the
  visual cortex by correlated neuronal activity. Science  \textbf{255}(5041),
  209--212 (1992)

\bibitem{mozafari2019bio}
Mozafari, M., Ganjtabesh, M., Nowzari-Dalini, A., Thorpe, S.J., Masquelier, T.:
  Bio-inspired digit recognition using reward-modulated spike-timing-dependent
  plasticity in deep convolutional networks. Pattern recognition  \textbf{94},
  87--95 (2019)

\bibitem{mozafari2018first}
Mozafari, M., Kheradpisheh, S.R., Masquelier, T., Nowzari-Dalini, A.,
  Ganjtabesh, M.: First-spike-based visual categorization using
  reward-modulated stdp. IEEE transactions on neural networks and learning
  systems  \textbf{29}(12),  6178--6190 (2018)

\bibitem{NEURIPS2019_9015}
Paszke, A., Gross, S., Massa, F., Lerer, A., Bradbury, J., Chanan, G., Killeen,
  T., Lin, Z., Gimelshein, N., Antiga, L., Desmaison, A., Kopf, A., Yang, E.,
  DeVito, Z., Raison, M., Tejani, A., Chilamkurthy, S., Steiner, B., Fang, L.,
  Bai, J., Chintala, S.: Pytorch: An imperative style, high-performance deep
  learning library. In: Wallach, H., Larochelle, H., Beygelzimer, A.,
  d\textquotesingle Alch\'{e}-Buc, F., Fox, E., Garnett, R. (eds.) Advances in
  Neural Information Processing Systems 32, pp. 8024--8035. Curran Associates,
  Inc. (2019),
  \url{http://papers.neurips.cc/paper/9015-pytorch-an-imperative-style-high-performance-deep-learning-library.pdf}

\bibitem{poggio2017and}
Poggio, T., Mhaskar, H., Rosasco, L., Miranda, B., Liao, Q.: Why and when can
  deep-but not shallow-networks avoid the curse of dimensionality: a review.
  International Journal of Automation and Computing  \textbf{14}(5),  503--519
  (2017)

\bibitem{pogodin2021towards}
Pogodin, R., Mehta, Y., Lillicrap, T., Latham, P.E.: Towards biologically
  plausible convolutional networks. Advances in Neural Information Processing
  Systems  \textbf{34},  13924--13936 (2021)

\bibitem{saunders2019locally}
Saunders, D.J., Patel, D., Hazan, H., Siegelmann, H.T., Kozma, R.: Locally
  connected spiking neural networks for unsupervised feature learning. Neural
  Networks  \textbf{119},  332--340 (2019)

\bibitem{saunders2018stdp}
Saunders, D.J., Siegelmann, H.T., Kozma, R., et~al.: Stdp learning of image
  patches with convolutional spiking neural networks. In: 2018 international
  joint conference on neural networks (IJCNN). pp.~1--7. IEEE (2018)

\bibitem{saunders2020minibatch}
Saunders, D.J., Sigrist, C., Chaney, K., Kozma, R., Siegelmann, H.T.: Minibatch
  processing for speed-up and scalability of spiking neural network simulation.
  In: 2020 International Joint Conference on Neural Networks (IJCNN). pp.~1--8.
  IEEE (2020)

\bibitem{schemmel2010wafer}
Schemmel, J., Br{\"u}derle, D., Gr{\"u}bl, A., Hock, M., Meier, K., Millner,
  S.: A wafer-scale neuromorphic hardware system for large-scale neural
  modeling. In: 2010 IEEE International Symposium on Circuits and Systems
  (ISCAS). pp. 1947--1950. IEEE (2010)

\bibitem{schrimpf2020brain}
Schrimpf, M., Kubilius, J., Hong, H., Majaj, N.J., Rajalingham, R., Issa, E.B.,
  Kar, K., Bashivan, P., Prescott-Roy, J., Geiger, F., et~al.: Brain-score:
  Which artificial neural network for object recognition is most brain-like?
  BioRxiv p. 407007 (2020)

\bibitem{schultz1997neural}
Schultz, W., Dayan, P., Montague, P.R.: A neural substrate of prediction and
  reward. Science  \textbf{275}(5306),  1593--1599 (1997)

\bibitem{mulitdigitmnist}
Sun, S.H.: Multi-digit mnist for few-shot learning (2019),
  \url{https://github.com/shaohua0116/MultiDigitMNIST}

\bibitem{sutton2018reinforcement}
Sutton, R.S., Barto, A.G.: Reinforcement learning: An introduction. MIT press
  (2018)

\bibitem{tavanaei2019deep}
Tavanaei, A., Ghodrati, M., Kheradpisheh, S.R., Masquelier, T., Maida, A.: Deep
  learning in spiking neural networks. Neural Networks  \textbf{111},  47--63
  (2019)

\bibitem{weidel2021unsupervised}
Weidel, P., Duarte, R., Morrison, A.: Unsupervised learning and clustered
  connectivity enhance reinforcement learning in spiking neural networks.
  Frontiers in computational neuroscience  \textbf{15}, ~18 (2021)

\end{thebibliography}

\section*{Supplementary Material}
\beginsupplement
\begin{table}[h]
\begin{center}
\centering
\caption{BioLCNet (hyper-)parameters for the MNIST task; best-performing value for (hyper-)parameters subject to grid search are in bold.}
\small
\begin{tabular}{|c|c|}
\hline
Parameter  & Value  \\ \hline
    $u_{thr_0}$   &  -52 $(mV)$    \\ \hline
    $u_{rest}, u_{reset}$      & -65 $(mV)$   \\ \hline
    $g_0$  &  0.05 $(mV)$\\ \hline
    $\tau_g$          &   $10^6$ $(ms)$\\ \hline
    $\Delta t_{ref}$      & 5 $(ms)$  \\ \hline
    $\tau_m$       & 20 $(ms)$  \\ \hline
    $f_{max}$        &  128 $(Hz)$ \\ \hline
    $h_{in}$, $w_{in}$&  22 \\ \hline
    $n_{out}$ &  [100, 500, \textbf{1000}] \\ \hline
    $ch_{lc}$        & [25, 50, \textbf{100}, 250]  \\ \hline
    $k$       & [11, 13, \textbf{15}, 17] \\ \hline
    $s$       &  [2, 3, \textbf{4}] \\ \hline 
    $T_{adapt}$, $T_{dec}$, $T_{learn}$        & 256 $(ms)$  \\ \hline
    $(\eta_{pre}, \eta_{post})_{STDP}$ &  $(0.0001, 0.01)$ \\ \hline
    $(\eta_{pre}, \eta_{post})_{R-STDP}$ &  $(0.1, 0.1)$ \\ \hline
    $\gamma$ &  1 \\ \hline
    $\eta_{rpe}$ &  [(static), 0.075, \textbf{0.125}, 0.175, 0.25] \\ \hline
    $\alpha$ &  0.9 \\ \hline
    $w_{inh}$    & -100  \\ \hline
    $c_{norm}$        &  0.25 \\ \hline
\end{tabular}
\label{tablesupp}
\end{center}
\end{table}

\section*{XOR MNIST}
\begin{figure}[htp]
    \centering
  \subfloat(a\label{1a}){%
      \includegraphics[width=0.2\textwidth]{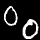}}
  \subfloat(b\label{1b}){%
        \includegraphics[width=0.2\textwidth]{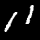}}
  \subfloat(c\label{1c}){%
        \includegraphics[width=0.2\textwidth]{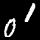}}
  \subfloat(d\label{1d}){%
        \includegraphics[width=0.2\textwidth]{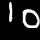}}
  \caption{Four samples of the XOR MNIST dataset; (a) 00 and (b) 11 represent class 0 (false), (c) 01 and (d) 10 represent class 1 (true).}
  \label{fig1_supp} 
\end{figure}

As the last experiment, we evaluated the performance of BioLCNet on a task defined on a portion of the multi-digit MNIST dataset \cite{mulitdigitmnist}. The multi-digit MNIST is created by putting the different MNIST samples together to create images with multiple digits, and the original dataset contains 100 classes (from 00 to 99). In this experiment, we extracted four classes, 00, 11, 01, 10 from this dataset and labeled each sample according to the XOR logic, i.e., samples representing 00 and 11 are given the label 0 (false output of an XOR logical gate), and samples representing 01 and 10 are given the label 1 (true output of an XOR logical gate). Each sample in this dataset is a 40 by 40 square image. Fig.~\ref{fig1_supp} represents four different samples of XOR MNIST. We created a balanced training and test set each including 10000 samples, and report the test accuracy on the whole test set averaged over ten independent runs. The best performance was achieved using 1000 channels in the LC layer, with a filter size of 32 and a stride of 4. The rewarding mechanism and the other related hyperparameters are set the same as the best-performing MNIST model. The final classification accuracy with this configuration was 84.35$\pm$1.27. This task is not a simple classification problem, as the R-STDP dynamics needs to first figure out how to group the four different patterns into two classes, and consequently reinforce the corresponding connections to the decoding layer. These should be done with a single scalar feedback signal (reward) only based on the validity of the network's decision. We used the same network architecture as the other experiments, however, adding more hidden layers to the network may be a way to enhance the performance on this task.

\end{document}